\documentclass[10pt]{article}
\usepackage[preprint]{tmlr}

\usepackage{hyperref}
\usepackage{url}

\usepackage{graphicx}
\graphicspath{{figures/}}
\DeclareGraphicsExtensions{.pdf,.png,.jpg}
\usepackage{subcaption}
\usepackage{booktabs}       %
\usepackage{makecell}
\usepackage{enumitem}

\usepackage{mathtools}
\usepackage{bm}
\usepackage{amsfonts}       %
\DeclareMathOperator*{\argmax}{arg\,max}

\newcommand{\MNAME}{Knockout}
\newcommand{\indep}{\mbox{${}\perp\mkern-11mu\perp{}$}}
\newcommand{\dep}{\mbox{${}\not\!\perp\mkern-11mu\perp{}$}}
\newcommand{\EE}{\mathbb{E}}
\newcommand{\II}{\mathbb{I}}
\newcommand{\PP}{\mathrm{P}}

\newcommand{\Mset}{\mathcal{M}}
\newcommand{\Mmask}{\bm{M}}
\newcommand{\DEF}{\bar{x}}
\newcommand{\GHI}{\dot{x}}

\title{\MNAME: A simple way to handle missing inputs}

\author{\name Minh Nguyen$^\dagger$ \email bn244@cornell.edu \\
      \addr Cornell University
      \AND
      \name Batuhan K. Karaman$^\dagger$ \email kbk46@cornell.edu \\
      \addr Cornell University
      \AND
      \name Heejong Kim$^\dagger$ \email hek4004@med.cornell.edu \\
      \addr Weill Cornell Medicine
      \AND
      \name Alan Q. Wang$^\dagger$ \email aw847@cornell.edu \\
      \addr Cornell University
      \AND
      \name Fengbei Liu$^\dagger$ \email fl453@cornell.edu \\
      \addr Cornell University
      \AND
      \name Mert R. Sabuncu \email msabuncu@cornell.edu\\
      \addr Cornell University and Weill Cornell Medicine\\
      for the Alzheimer’s Disease Neuroimaging Initiative~\thanks{
Data used in preparation of this article were obtained from the Alzheimer’s Disease Neuroimaging Initiative
(ADNI) database (adni.loni.usc.edu). As such, the investigators within the ADNI contributed to the design
and implementation of ADNI and/or provided data but did not participate in analysis or writing of this report.
A complete listing of ADNI investigators can be found at: http://adni.loni.usc.edu/wp-content/uploads/how\_to\_apply/ADNI\_Acknowledgement\_List.pdf
\\ $^\dagger$: Equal contribution
}}

\def\month{07}
\def\year{2025}
\def\openreview{\url{https://openreview.net/forum?id=K71y5pge84}}

\begin{document}

\maketitle

\begin{abstract}
Deep learning models benefit from rich (e.g., multi-modal) input features.
However, multi-modal models might be challenging to deploy, because some inputs may be missing at inference.
Current popular solutions include marginalization, imputation, and training multiple models.
Marginalization achieves calibrated predictions, but it is computationally expensive and only feasible for low dimensional inputs.
Imputation may result in inaccurate predictions, particularly when high-dimensional data, such as images, are missing.
Training multiple models, where each model is designed to handle different subsets of inputs, can work well but requires prior knowledge of missing input patterns.
Furthermore, training and retaining multiple models can be costly.
We propose an efficient method to learn both the conditional distribution using full inputs and the marginal distributions.
Our method, \MNAME, randomly replaces input features with appropriate placeholder values during training.
We provide a theoretical justification for \MNAME~and show that it can be interpreted as an implicit marginalization strategy.
We evaluate \MNAME~across a wide range of simulations and real-world datasets and show that it offers strong empirical performance.
\end{abstract}

\section{Introduction}
In many real-world applications of machine learning (ML) and statistics, not all variables might be available for every data point.
This issue, also known as missingness, is well-studied in the literature~\citep{little2019statistical} and common in fields like healthcare, social sciences, and environmental studies.
From a Bayesian perspective, missingness can be solved by marginalization, where we would like a model to marginalize out the missing variables from the conditioning set.
However, during training, we often do not know which features will be missing at inference.

In lieu of training multiple models for every missingness pattern, a common strategy is imputation, which uses a point estimate (usually the mean or mode or a constant) to impute the missing features~\citep{le2020linear}.
This can be seen as approximating the marginalization with a delta function.
More sophisticated imputation methods use EM~\citep{josse2019consistency} or neural-based networks~\citep{mattei2019miwae,ipsen2022deal}.
Although many prior methods may work well in some instances, they may not scale readily to high-dimensional inputs like images~\citep{kyono2021miracle,you2020handling}, require additional networks for generation of missing variables~\citep{ipsen2022deal}, only apply to continuous inputs~\citep{le2020neumiss,le2021sa}, assume linearity of predictors~\citep{le2020linear}, or make assumptions about the data distribution~\citep{hazan2015classification}.

In this work, we propose a simple, effective, and theoretically-justified augmentation strategy, called \MNAME, for handling missing inputs.
During training, features are randomly ``knocked out'' and replaced by constant ``placeholder'' values.
At inference time, using the placeholder value corresponds mathematically to estimation with the appropriate marginal distribution.
\MNAME~can be seen as implicitly maximizing the likelihood of a weighted sum of the conditional estimators and all desired marginals \textit{in a single model}.
We demonstrate the broad applicability of \MNAME~in a suite of experiments using synthetic and real-world data from multiple modalities (image-based and tabular).
Real world experiments include Alzheimer's forecasting, noisy label learning, multi-modal MR image segmentation, multi-modal image classification, and multi-view tree genus classification.
We show the effectiveness of \MNAME~in handling low and high-dimensional missing inputs compared against competitive baselines.

\section{Method}\label{sec:method}

\subsection{Background}\label{ssec:background}
The goal of supervised ML is to learn the conditional distribution $p(Y|\bm{X})$ where $Y$ is the output (predictive target) and $\bm{X}\in \mathbb{R}^N$ is the vector of inputs or features.
The prediction for a new sample $\bm{x}$ is $\hat{y} = \argmax_{Y} p(Y|\bm{X}=\bm{x})$.
However, in many practical applications, not all features may be present.
When a feature $X_i$ is missing, the vector $\bm{X}_{-i}$ denotes the non-missing features.
In general, multiple features may be missing at a time.
We can represent this with a missingness indicator set $\Mset$ and corresponding non-missing features as $\bm{X}_{-\Mset}$.
In this case, what we really want is $p(Y|\bm{X}_{-\Mset})$.
How can we account for missingness?

A simple approach is to train a separate model for $p(Y|\bm{X}_{-\Mset})$, i.e.~a model that takes only the non-missing features $\bm{X}_{-\Mset}$ as inputs.
However, this is expensive because a separate model is needed for each missingness pattern.
Furthermore, there is no sharing of information between these separate models, even though they are theoretically related.

Another approach rewrites $p(Y|\bm{X}_{-\Mset})$ using the already available $p(Y|\bm{X})$:
\begin{equation}
    p(Y|\bm{X}_{-\Mset}) = \int p(Y, \bm{X}_\Mset| \bm{X}_{-\Mset}) d\bm{X}_\Mset = \int p(Y|\bm{X}) p(\bm{X}_\Mset|\bm{X}_{-\Mset}) d\bm{X}_\Mset. \label{eq:marginalization}
\end{equation}
The goal now is to obtain $p(\bm{X}_\Mset | \bm{X}_{-\Mset})$ and perform the integration over all possible $\bm{X}_\Mset$.
Imputation methods approximate Eq.~\eqref{eq:marginalization} by replacing $p(\bm{X}_\Mset | \bm{X}_{-\Mset})$ with a delta function.
For example, ``mean imputation'' uses the mean of the missing features $\bm{X}_\Mset$, i.e. $\EE[\bm{X}_\Mset]$, for $\bm{X}_\Mset$ itself.
In Eq.~\ref{eq:marginalization}, this corresponds to approximating $p(\bm{X}_\Mset|\bm{X}_{-\Mset}) \approx \delta(\EE[\bm{X}_\Mset])$, a delta function. 
While convenient and commonly used, mean imputation ignores the dependency between $\bm{X}_\Mset$ and $\bm{X}_{-\Mset}$, and does not account for any uncertainty.

More sophisticated imputation methods capture the interdependencies between inputs~\citep{troyanskaya2001missing,stekhoven2012missforest}, for example by explicitly modeling $p(\bm{X}_\Mset | \bm{X}_{-\Mset})$ by training a separate model. 
The point estimate $\bm{x}_\Mset = \argmax_{\bm{X}_\Mset} p(\bm{X}_\Mset|\bm{X}_{-\Mset})$ can be used at inference time for the missing $\bm{X}_\Mset$.
While properly accounting for interdependencies between inputs, this approach requires fitting a separate model for $p(\bm{X}_\Mset|\bm{X}_{-\Mset})$.
In multiple imputation, multiple samples from $p(\bm{X}_\Mset|\bm{X}_{-\Mset})$ are drawn and a Monte Carlo approximation is used to estimate the integral on the RHS of Eq.~\ref{eq:marginalization}~\citep{mattei2019miwae,kyono2021miracle,ipsen2022deal}.
Although this is more accurate than single imputation, it is not effective in high dimensional space.

\subsection{\MNAME}\label{ssec:Knockout}
We propose a simple augmentation strategy for neural network training called \MNAME~that enables estimation of the conditional distribution $p(Y|\bm{X})$ and all desired marginals $p(Y | \bm{X}_{-\mathcal{M}})$ in a single, high capacity, nonlinear model, such as a deep neural network.
During training, features are augmented by randomly ``knocking out'' and replacing them with constant, ``placeholder'' values.
At inference time, using the placeholder value corresponds mathematically to estimation with the suitable marginal distribution.

Specifically, let $\Mmask=[M_1,M_2,\dots, M_N] \in \{0, 1\}^N$ denote a binary, induced missingness indicator vector.
Let $\bm{\DEF}:=[\DEF_1,\DEF_2,\dots, \DEF_N]\in \mathbb{R}^N$ denote a vector of placeholder values.
Then, define $\bm{X}'(\Mmask, \bm{X})= \Mmask \odot \bm{\DEF} + (\bm{1}-\Mmask)\odot \bm{X}$ as augmented \MNAME~inputs,
where $\bm{1}$ is a vector of ones and $\odot$ denotes element-wise multiplication.
During one training iteration, a different \MNAME~input is used corresponding to a different randomly sampled $\Mmask$ for every data sample.
The model weights are trained to minimize the loss function with respect to $Y$, as is done regularly.

Two mild conditions are required to ensure proper training.
First, the placeholder values must be ``appropriate,'' as we will elaborate below.
For our theoretical treatment, we will use out-of support values as appropriate; i.e.~$\bm{\DEF}_\Mset \not\in \text{Support}(\bm{X}_\Mset)$.
Second, $\Mmask$ must be independent of $\bm{X}$ and $Y$, i.e. $\Mmask \indep \bm{X}, Y$.%
\footnote{Note it is not necessary that $M_i \indep M_j$ for any $i, j$.}
It follows straightforwardly that these two conditions lead to modeling the desired conditional and marginal distributions simultaneously.
First, since $\bm{\DEF}_\Mset$ is not in the support of $\bm{X}_\Mset$,
\begin{align}
    &\bm{X}_\Mset'=\bm{\DEF}_\Mset \Leftrightarrow \Mmask_\Mset = \bm{1},
    &\bm{X}_\Mset'\neq \bm{\DEF}_\Mset \Leftrightarrow \Mmask_\Mset=\bm{0} \ \text{and} \ \bm{X}_\Mset'=\bm{X}_\Mset,
\end{align}
where $\bm{0}$ and $\bm{1}$ are vectors of zeros and ones of appropriate shape.
Second, since $\bm{M}$ is independent of $\bm{X}$ and $Y$, it follows that imputing with the default value $\bm{\DEF}_\Mset$ is equivalent to marginalization of the missing variables defined by $\Mset$:
\begin{align}
    p(Y|\bm{X}_\Mset'{=}\bm{\DEF}_\Mset,\bm{X}_{-\Mset}'{=}\bm{x}_{-\Mset}) &= p(Y|\Mmask_\Mset{=}\bm{1},\Mmask_{-\Mset}{=}\bm{0},\bm{X}_{-\Mset}{=}\bm{x}_{-\Mset}) \\
    &= p(Y|\bm{X}_{-\Mset}{=}\bm{x}_{-\Mset}).
\end{align}
At the two extremes, no \MNAME{} ($\Mmask = \bm{0}$) corresponds to the original conditional distribution, and full \MNAME{} ($\Mmask = \bm{1}$) to the full marginal:
\begin{align}
    &p(Y|\bm{X}'{=}\bm{x}) = p(Y|\Mmask{=}\bm{0},\bm{X}{=}\bm{x}) = p(Y|\bm{X}{=}\bm{x}) \label{eq:first_in_set}, \\
    &p(Y|\bm{X}'{=}\bm{\DEF}) = p(Y|\Mmask{=}\bm{1}) = p(Y) . \label{eq:last_in_set}
\end{align}

For a new test input $\bm{x}$, the prediction when $\bm{x}_\Mset$ is missing is simply 
\begin{equation}
    \argmax_{Y} p(Y|\bm{X}_{-\Mset}{=}\bm{x}_{-\Mset}) = \argmax_{Y} p(Y|\bm{X}_\Mset'{=}\bm{\DEF}_\Mset,\bm{X}_{-\Mset}'{=}\bm{x}_{-\Mset}),
\end{equation}
i.e., the learned estimator with the augmented \MNAME~input. 

\subsubsection{\MNAME~as an Implicit Multi-task Objective}
The missingness indicator $\Mmask$ determines how inputs are replaced with appropriate placeholder values during training.
To satisfy the independence condition of $\Mmask$ with $\bm{X}$ and $\bm{Y}$, the variables $\Mmask$ are sampled independently from a distribution $p(\Mmask)$ during training.
We show that this training strategy can be viewed as a multi-task objective~\citep{caruana1997multitask} decomposed as a weighted sum of terms, where each term is a separate marginal weighted by the distribution of $\Mmask$.
Let $\ell$ denote the loss function to be minimized (e.g., mean-squared-error or cross-entropy loss):

\begin{align}
    L(\theta) &= \EE_{\bm{X}',Y} \ \ell(Y;f_\theta(\bm{X}'(\Mmask, \bm{X})) \\
    &= \EE_{\bm{X},Y}\EE_{\Mmask}\sum_{\bm{m} \in \Mmask}\II(\Mmask{=}\bm{m}) \ \ell(Y;f_\theta(\bm{X}'(\Mmask, \bm{X}))) \\
    &= \EE_{\bm{X},Y} \sum_{\bm{m}\in\Mmask} p(\Mmask{=}\bm{m}) \ \ell(Y;f_\theta(\bm{X}'(\bm{m}, \bm{X}))) \\
    &= \sum_{\bm{m}} p(\Mmask{=}\bm{m}) \ \EE_{\bm{X},Y} \ell(Y;f_\theta(\bm{X}'(\bm{m}, \bm{X}))),
\end{align}
where $\II$ is the indicator function.

If there is knowledge about the missingness patterns at inference (e.g., some $X_i$ and $X_j$ exhibit correlated missingness), one can design $p(\Mmask)$ appropriately to cover all the expected missing patterns, i.e.~by sampling $\bm{m}$ during training with different weights.
In the absence of such knowledge, the most general distribution for $\Mmask$ is i.i.d. Bernoulli.
A common way correlated missingness arises in real-world applications is in structured inputs like latent features or images, where the entire feature vector or whole image is missing. 
In our experiments, we demonstrate the superiority of \textit{structured} \MNAME{}, over naive i.i.d. \MNAME{}, when such correlated missingness is known a priori. 

\subsection{Choosing Appropriate Placeholder Values}\label{ssec:masking_values}
Our theoretical analysis assumes that the placeholder value $\DEF_i$ lies outside the support of $X_i$ (see Appendix~\ref{app:counterexample} for further discussion).
Previous work by~\cite{josse2019consistency} and~\cite{bertsimas2024simple} has shown that, with infinite data, mean imputation (for continuous variables) and out-of-support imputation (for both continuous and discrete variables) can be optimal.
However, in practice, data is finite, and empirical performance becomes more important.
In such cases, using out-of-support values as placeholders is often preferable and can be mathematically justified—particularly when $X_i$ is low-dimensional.
That said, for high-dimensional inputs such as vectors or images, out-of-range placeholders may lead to practical issues like unstable gradients or limited model capacity, making them less effective.
In the sections that follow, we relax the out-of-support assumption and offer practical recommendations for choosing placeholder values for different types of $X_i$, based on these considerations.

\subsubsection{Non-structured}
In this section, we recommend suitable placeholder values for non-structured, scalar-valued inputs.
\begin{itemize}
    \item \textbf{Categorical variable}: $\DEF_i$ can be $N_{X_i}+1$ if $X_i$ are integer-valued classes from 1 to $N_{X_i}$.
If one-hot encoded, $\DEF_i$ can be a vector of 0s.
    \item \textbf{Continuous variable and Non-empty Infeasible Set}: we can scale $X_i$ to $[0, 1]$ and choose $\DEF_i = -1$.
More generally, if $X_i$ has unbounded range but a non-empty infeasible set, then $\DEF_i$ can be set to a value in the infeasible set.
For example, if $X_i$ only takes positive values, then we can set $\DEF_i=-1$.
    \item \textbf{Continuous variable and Empty Infeasible Set}: we suggest applying $Z$-score normalization and choosing $\DEF_i$ such that it lies in a low probability region of the normalized $X_i$ such that $p(X_i{=}\DEF_i)\approx 0$.
As we argue in Appendix~\ref{app:cont_unbounded_proof}, this approach leads to an approximation of the desired marginal.
\end{itemize}

\begin{figure}[t]
\centering
\includegraphics[width=\linewidth]{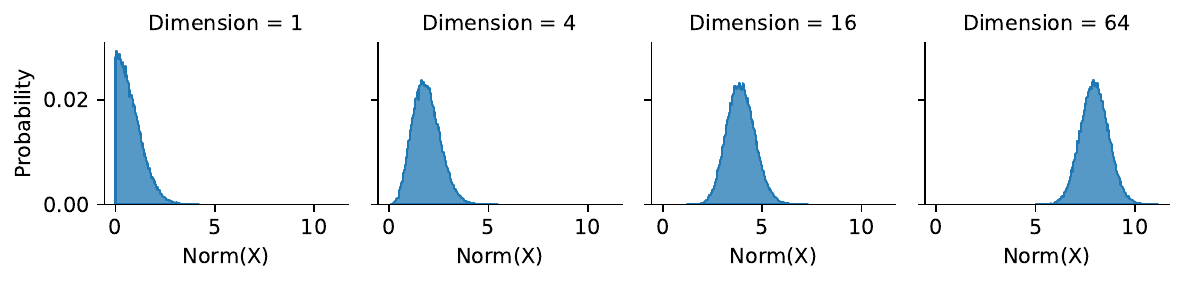}
\caption{The region of high density of standard Gaussian shifts away from the origin as the number of dimensions increases.
This motivates different choices of placeholder values at different dimensions.}\label{fig:where_is_z0}
\end{figure}

Although different distributions have different regions of low probability, we use the behavior of the Gaussian distribution as a guide to choose $\DEF_i$. 
Fig.~\ref{fig:where_is_z0} shows the histogram of the norm of points sampled from standard Gaussian distributions with different dimensionality.
For a univariate standard Gaussian, most of the points lie close to the origin so we should choose $\DEF_i$ far away from the origin.
However, as the dimension increases, most of the points lie on the hyper-sphere away from the origin so we should choose $\DEF_i$ to be the point at origin (i.e.~a vector of zeros).
Table~\ref{tbl:choose_z0} summarizes the choices of $\DEF_i$ for different types of inputs $X_i$.
\begin{table}[t]
  \caption{List of different types of $X_i$ and the recommended $\DEF_i$}\label{tbl:choose_z0}
  \centering
  \begin{tabular}{lccccc}
    \toprule
      \textit{Type of} $X_i$ & \textit{Example} & \textit{Dimension} & \textit{Support} & \textit{Normalized?} & $\bm{\DEF}_i$  \\
    \midrule
      Categorical  & Gender  & 1   & $\{1,\dots,N_{X_i}\}$    & N/A               & $N_{X_i} {+} 1$   \\
      Continuous    & Test scores & 1   & $[a, b]$              & Scale to [0, 1]   & -1              \\  
      Continuous    & Temperature & 1   & $[a, \infty)$ or $(-\infty, b]$          & Scale to $[0, \infty)$           & -1         \\  
      Continuous    & White noise & 1   & $(-\infty, \infty)$            & Z-score           & $\pm 10$         \\  
    \midrule
      Structured    & Images & $>$1000  & $[a, b]$              & Scale to [0, 1]   & $\bm{0}$    \\  
      Structured    & Latent vectors & $>$16 & $(-\infty, \infty)$            & Z-score           & $\bm{0}$     \\  
    \bottomrule
  \end{tabular}
\end{table}

\subsubsection{Structured}
For structured inputs like images and feature vectors, there are multiple choices of placeholder.
However, we found that \MNAME~applied with some out-of-support placeholder like $\bm{-1}$, can cause issues like unstable gradients.
Therefore, we recommend an appropriate placeholder to be either the image of all 0s or the mean image.
When $Z$-score normalization is applied, the 0 image and the mean image coincide.
Theoretically, it is well known that the mean of a high dimensional random variable, such as a Gaussian, has very low probability~\citep{vershynin2018high} (also see Fig.~\ref{fig:where_is_z0}).
We believe this recommendation balances the tension between ensuring an extremely-low probability placeholder with proper convergence and performance.
For an empirical demonstration, see Appendix~\ref{sec:analysis_placeholders_structured_knockout}.

\subsection{Observed Missingness during Training}\label{ssec:observed_missingness}
The treatment above assumes complete training data, and inference-time missingness only.
We now consider the situation where training data has \textit{observed} missing inputs.
Let $\bm{N}$ be the binary mask indicating the observed data missingness.
$\bm{N}$ is different from $\bm{M}$, which denotes the missingness induced by \MNAME~during training.
Thus, $\bm{N}$ is fixed for a data sample, while $\bm{M}$ is stochastic.
Observed missingness generally falls under the following scenarios~\citep{little2019statistical}.

\textbf{Missing Completely at Random (MCAR):}
This implies that $\bm{N} \indep \bm{X}, Y$.
Let $\bm{M}':= \bm{N} \lor \bm{M}$ be the augmented masking indicator, where $\lor$ denotes the logical OR operation.
Since $\bm{N} \indep \bm{X}, Y$ and $\bm{M} \indep \bm{X}, Y$, so $\bm{M}' \indep \bm{X}, Y$.
Therefore, we can obtain the same result in Section~\ref{ssec:Knockout} when using $\bm{M}'$ instead of $\bm{M}$ as the masking indicator vector.
This implies that \MNAME~can be applied to MCAR training data simply by masking all the missing values using the same placeholders $\bm{\DEF}$.

\textbf{Missing at Random (MAR) and Missing not at Random (MNAR):}
This implies that $\bm{N} \dep \bm{X}, Y$.
Thus, we cannot replace the missing values in training data using the same placeholders.
However, we can substitute these values using placeholders that are different from $\bm{\DEF}$ but are also outside the support of the input variables (or very unlikely values).
Let the placeholders for the data missingness be $\bm{\GHI}\neq\bm{\DEF}$.
During training, \MNAME~still randomly masks out input variables, including those that are not observed in the data.
Thus, the results in Section~\ref{ssec:Knockout} still hold since $\bm{M} \indep \bm{X}, Y$.

During inference, if we know a priori that $x_i$ of a sample is missing not at random, then we can use $\GHI_i$ as the placeholder.
Otherwise, if we know $x_i$ is missing completely at random, we use $\DEF_i$.
Unless stated otherwise, \MNAME~always uses a different placeholder for MAR and MNAR variables.

\begin{table}[t]
\caption{Summary of experimental setups}\label{tbl:summary_of_experiments}
\centering
\scalebox{0.9}{
\begin{tabular}{llcccc}
\toprule
\textit{Task}             & \textit{Type of} $X_i$ & \textit{Dimension} & \textit{Normalized?}   & $\bm{\DEF_i}$      & $\bm{\GHI}$ \\ \midrule
Simulations               & Categorical/Continuous & 1                  & Z-score (Cont. $X_i$) &  10 &    -10       \\
Alzheimer's Forecasting   & Continuous             & 1                  & Z-score        & 10       & -10       \\
Privileged Information    & Continuous             & 1                  & Scale to [0, 1] & -1            & N/A        \\
Tumor Segmentation        & Structured (images)    & $256^3$            & Scale to [0, 1]        & $\bm{0}$      & N/A        \\
Tree Genus Classification & Structured (latent)    & $768$ or $2048$              & Z-score                & $\bm{0}$      & N/A          \\
Prostate Cancer Detection & Structured (latent)    & $256$              & Z-score                & $\bm{0}$      & N/A          \\
Food Classification       & Structured (latent)    & $768$              & Z-score                & $\bm{0}$      & N/A          \\
\bottomrule
\end{tabular}}
\end{table}

\section{Related Work}
\MNAME~is inspired by other methods with different objectives.
Dropout~\citep{srivastava2014dropout,gal2016dropout} prevents overfitting by randomly dropping units (hidden and visible) during training, effectively marginalizing over model parameters.
During inference, this marginalization can be approximated by predicting once without dropout~\citep{srivastava2014dropout} or averaging multiple predictions with dropout~\citep{gal2016dropout}.
Blankout~\citep{maaten2013learning} and mDAE~\citep{chen2014marginalized} learn to marginalize out the effects of corruption over inputs.
In contrast, \MNAME~learns different marginals to handle varied missing input patterns.

Imputation techniques impute missing inputs explicitly, e.g., using mean, median, or mode values. 
Model-based imputation methods predict missing values via k-nearest neighbors~\citep{troyanskaya2001missing}, chained equations~\citep{van2011mice}, random forests~\citep{stekhoven2012missforest}, autoencoders~\citep{gondara2018mida,ivanov2018variational,lall2022midas}, RNN~\citep{nguyen2020predicting}, GANs~\citep{yoon2018gain,li2018misgan,belghazi2019learning}, or normalizing flows~\citep{li2020acflow}.
Although more accurate than simple imputation, model-based imputation incurs significant additional computation costs, especially with high-dimensional inputs.
Some approaches~\citep{ma2021smil,peis2022missing} require additional training of multiple VAEs or sub-networks.
Other approaches~\citep{mattei2019miwae,ma2019eddi} require training only one VAE but assume homogeneous data (all continuous variables or all binary variables), limiting flexibility.
In contrast, \MNAME~uses a single classifier, avoiding the need for separate models to impute missing inputs explicitly.

Another relevant line of work is causal discovery~\citep{spirtes2000causation}, which often involves fitting a model using different subsets of available inputs and multiple distributions simultaneously~\citep{lippe2021efficient,james2023participatory}.
To reduce computational cost, it is common to train a single model that can handle different subsets of inputs using dropout~\citep{brouillard2020differentiable,lippe2021efficient,ke2023neural,nguyen2024fasticp,nguyen2024glacial}.

Techniques like \MNAME~are often used in practice to train a single neural network that models multiple distributions, but are often justified empirically with little care taken in choosing placeholder values.
Many works use zeros without theoretical justification~\citep{belghazi2019learning,ke2023neural,brouillard2020differentiable,lippe2021efficient}.
GAIN~\citep{yoon2018gain} and MisGAN~\citep{li2018misgan} impute using out-of-support values similar to \MNAME.
However, both are limited in their treatment by assuming that the supports are bounded, and do not consider categorical variables.
While the approach is similar to some prior work for structural inputs~\citep{neverova2015moddrop,parthasarathy2020training} or low-dimensional inputs~\citep{bertsimas2024simple}, \MNAME's theoretical backing shows that it can handle multiple data types and multiple missingness types (complete/MCAR/MAR/MNAR).
Similarly, methods like Selective MIM~\citep{van2023missing} have been proposed to enhance model performance in high-dimensional settings by selectively encoding informative missing patterns.
Many self-supervised learning techniques can be interpreted as training to reconstruct the inputs with \MNAME.
In addition, \MNAME~can be trained with standard empirical risk minimization while some approaches need more complex optimization~\citep{ma2021smil,ma2022multimodal}.
For example, masked language modeling~\citep{devlin2019bert} randomly maps tokens to an unseen ``masked'' token.
Denoising autoencoders~\citep{vincent2010stacked} randomly replace image patches with black patches, which are arguably out of the support of natural images.

\section{Experiments}
In all experiments, unless stated otherwise, we compare \MNAME~against a \textbf{common baseline} model trained on complete data, which, at inference time, imputes missing variables with mean (if continuous) or mode (if discrete) values.
If the training is done on incomplete data with observed missing variables, imputed with mean/mode, we denote this as \textbf{common baseline*}.
For most results we report a variant of \MNAME~but with sub-optimal placeholders (i.e.~mean/mode for continuous/categorical features).
We denote this variant as \textbf{\MNAME*}.
Note that both \MNAME* and common baseline* use the same placeholder (mean/mode), with the only difference being that \MNAME*-trained models observe randomly knocked-out missingness \textit{in addition to} (possible) observed missingness during training.

In all \MNAME~implementations, we choose random knockout rates such that, in expectation, half of the mini-batches have no induced missing variable.
If $d$ is the number of input variables and $r$ is the knockout rate, we choose $r$ such that the probability of not knocking out any variables in a mini-batch, i.e. ${(1 - r)}^d$ is approximately 0.5.
In batches with induced missingness, variables (or groups of variables in structured \MNAME) are independently removed, with a probability equal to the knockout rate.
The summary of the experimental setups are shown in Table~\ref{tbl:summary_of_experiments}.

\begin{figure}[t]
\centering
\includegraphics[width=\linewidth]{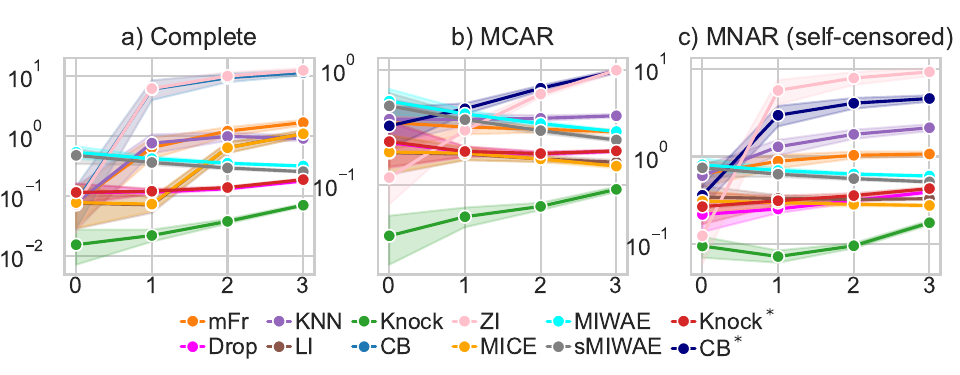}
    \caption{Test MSE evaluated against Bayes optimal prediction ($\EE[Y|\bm{X}]$) averaged over 10 repetitions (lower is better).
    X axis indicates the number of missing variables at inference time.
    mFr: missForest, Drop: dropout, Knock: \MNAME, CB: common baseline, ZI: zero-imputation with mask, sMIWAE: supMIWAE. KNN: k-nearest-neighbor imputation, LI: linear regression imputation}\label{fig:sim1}
\end{figure}

\subsection{Simulations}
We perform regression simulations to predict the output $Y$ from the input $\bm{X}\in\mathbb{R}^9$.
In each simulation, we sample 30k data points in total and use 10\% for training.
Since the input is low-dimensional, we can additionally compare against sophisticated but computationally expensive imputation methods. The baselines for the simulation are:
\begin{itemize}[itemsep=0pt, topsep=0pt]
    \item Inpute Dropout~\citep{srivastava2014dropout}: Randomly sets a subset of input features to zero during training.

    \item Zero Impute with Indicator: Replaces missing values with zeros and adds a binary indicator variable to denote the presence of missing data.

    \item MICE~\citep{van2011mice}: Iteratively imputes missing values by modeling each variable with missing data as a function of other variables in a round-robin fashion.

    \item missForest~\citep{stekhoven2012missforest}: Uses random forests to iteratively predict and fill in missing values.

    \item MIWAE~\cite{mattei2019miwae}: Imputes missing data using a VAE trained with importance-weighted bounds.

    \item supMIWAE~\citep{ipsen2022deal}: A supervised extension of MIWAE.

    \item KNN Impute: Fills in missing values by averaging the values of the k-nearest neighbors.

    \item Linear Regression Impute: Imputes missing values using predictions from a linear regression model.
\end{itemize}
All methods use the same neural network architecture composed of a 3-layer multi-layer perceptron (MLP) with hidden layers 100 and ReLU activations.
Training is done using Adam~\citep{kingma2015adam} with learning rate 3e-3 for 5k steps.

We generate training data corresponding to complete training data, MCAR training data, and MNAR training data.
For MCAR data, each feature is independently missing with a probability of 10\%.
For MNAR data, we adopt the self-censored missing setup where a variable is missing if its value is above the variable 90th percentile.
To ensure that performance comparisons reflect only differences in the training data, we keep the test data consistent across experiments.
Specifically, during testing, we apply a predefined missingness pattern (i.e., a specific set of missing variables) uniformly to all samples in the test set.
We repeat this procedure exhaustively for all missingness patterns involving up to 3 missing variables.
This resulted in 130 different missing patterns.

We evaluate the models' predictions against the MMSE-minimizing Bayes optimal predictions: $\EE[Y | \bm{X}]$ (please see Appendix~\ref{app:regression} for evaluation against observations $Y$).
Fig.~\ref{fig:sim1} shows the results of 10 repetitions of this simulation.
\MNAME~outperforms all the baselines regardless of the types of training data (complete, MCAR, or MNAR).
Dropout and \MNAME* achieve similar performance but both are worse than \MNAME; this underscores the importance of choosing an appropriate placeholder value.
We also experimented with varying the training-test split size to simulate low-data and data-rich regimes in Appendix~\ref{app:regression}.

\begin{figure}[ht]
\centering
\includegraphics[width=.9\linewidth]{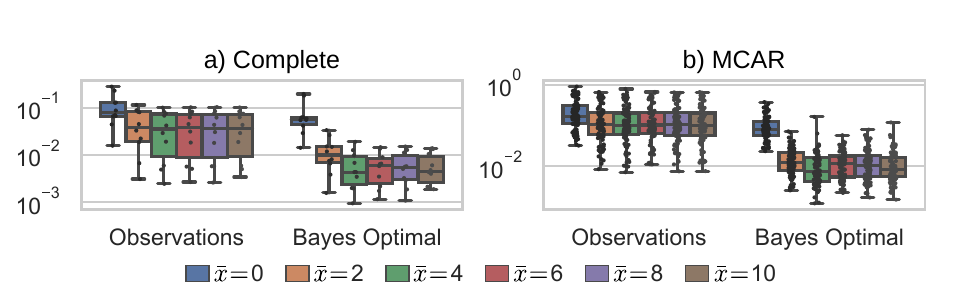}
    \caption{Lower test MSE in the regression simulation is achieved by selecting placeholder value that is far from the mean (i.e. 0).
    Observations: evaluate against $Y$. Bayes Optimal: evaluate against $\EE[Y | \bm{X}]$.
    }\label{fig:abl_z0}
\end{figure}

In addition, we perform an ablation to verify that choosing appropriate placeholders is critical for getting good performance (see Fig.~\ref{fig:abl_z0}).
In this ablation, the value of the placeholder of \MNAME~is selected from the set \{0, 2, 4, 6, 8, 10\}. 
After feature normalization, the mean value of a feature is 0 and using the placeholder value of 0 resulting in bad prediction performance (higher MSE).
Choosing placeholder value that is away from 0 (i.e.~away from the feature mean) lowers the test MSE in both complete data (Fig.~\ref{fig:abl_z0}a) and missing data (Fig.~\ref{fig:abl_z0}b) setup.
We also include an comparison aganist the common baseline with $\DEF$ imputation in Appendix~\ref{app:bin_classification} to illustrate the benefit of \MNAME~training strategy.
This baseline is similar to \MNAME~during inference but does not employ the augmentation strategy of \MNAME~during training.

To support our proposal for handling NMAR data, we include an ablation study comparing \MNAME~against its ablated version named \MNAME-.
While \MNAME~follows the procedure outlined in Section~\ref{ssec:observed_missingness}, 
\MNAME- treats all observed data uniformly, without distinguishing NMAR observations.
Both methods were trained using the same MNAR data detailed previously.
The results of this comparison are presented in Table~\ref{tab:mnar_abl}, which reports the MSE across varying levels of missingness.
\MNAME~consistently achieves lower MSEs, demonstrating the benefit of explicitly accounting for NMAR mechanisms.

\begin{table}[h]
\centering
\begin{tabular}{lllll}
\toprule
\textbf{Method} & \textbf{fm = 0} & \textbf{fm = 1} & \textbf{fm = 2} & \textbf{fm = 3} \\
\midrule
CB       & 0.361 $\pm$ 0.183 & 2.953 $\pm$ 3.870 & 4.053 $\pm$ 5.616 & 4.585 $\pm$ 6.109 \\
\MNAME-  & 0.144 $\pm$ 0.039 & 0.089 $\pm$ 0.055 & 0.107 $\pm$ 0.071 & 0.163 $\pm$ 0.141 \\
\MNAME   & 0.095 $\pm$ 0.044 & 0.072 $\pm$ 0.058 & 0.095 $\pm$ 0.062 & 0.175 $\pm$ 0.156 \\
\bottomrule
\end{tabular}
\caption{\MNAME's vs. \MNAME-'s MSE across varying numbers of missing features (\textbf{fm}) during inference.}\label{tab:mnar_abl}
\end{table}

\subsection{Missing Clinical Variables in Alzheimer's Disease Forecasting}
We demonstrate \MNAME's effectiveness in handling observed missing data during a real-world clinical task: predicting whether patients with mild cognitive impairment will progress to Alzheimer’s Disease within the next five years.
We use data from the Alzheimer's Disease Neuroimaging Initiative (ADNI) database~\citep{mueller_2005_ways} and adopt the state-of-the-art model from~\citep{karaman_2022_machine}.
Input features $\bm{X}$ include subject demographics, genetics, and clinical/imaging biomarkers.
The target $Y$ is a binary vector and indicates AD diagnosis in each of the five follow-up years.
For \MNAME, we use an out-of-range value of 10 for induced missingness and -10 for observed missingness, both during training and testing.
Further details about the dataset and experimental setup are provided in Appendix~\ref{app:ad}.
Fig.~\ref{fig:ad} shows the average AUROC scores when individual input features are omitted during inference.
These results are averaged over 10 random 80-20 train-test splits.
For each split, we compute the AUROC by averaging AUROC scores across the five follow-up years.
\MNAME~outperforms both the common baseline and \MNAME* in the vast majority of cases,highlighting the importance of selecting an effective placeholder strategy.

\begin{figure}[t]
\centering
\includegraphics[width=\linewidth]{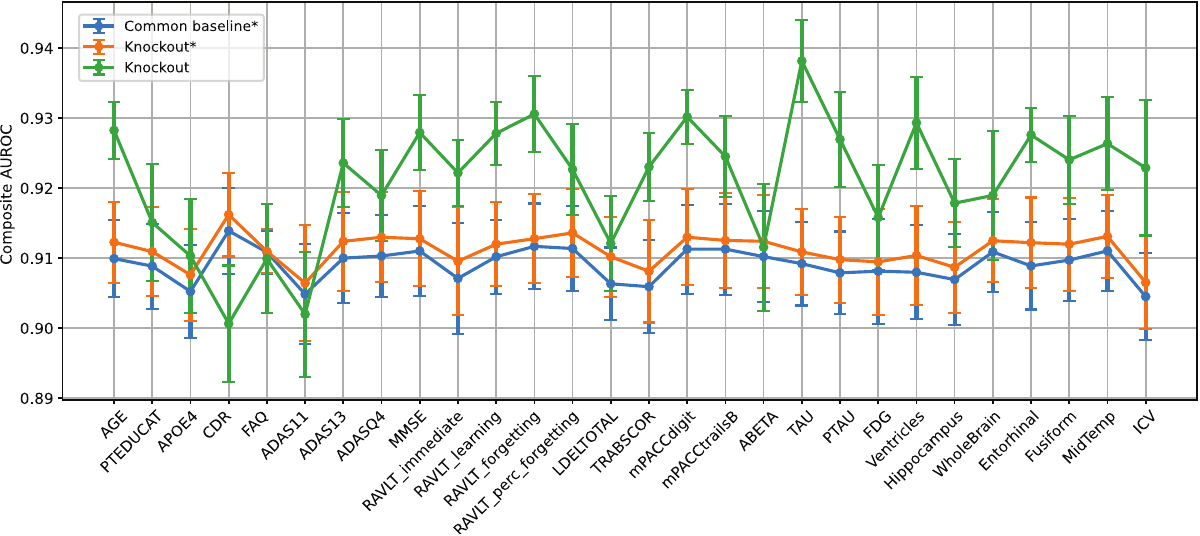}
\caption{AUROC scores obtained for the three model variants when each input feature is missing during inference (x-axis) in the Alzheimer's Disease forecasting experiment. Error bars indicate the standard error across 10 train-test splits.}\label{fig:ad} 
\end{figure}

\subsection{Privileged Information for Noisy Label Learning}
\MNAME~can be used to incorporate privileged information (PI) into the learning process.
For instance, PI might include demographic details that may be partially or completely missing during inference \citep{nguyen2024adapting}.
We focus on a noisy label learning task, where the objective is to leverage additional PI—such as the annotator’s ID or the time of annotation—to improve the model's robustness against label noise.
Due to the absence of PI in testing, existing methods~\citep{ortiz2023does,wang2023pi} require an auxiliary classification head for PI utilization.
\MNAME~can be directly applied with a method that accepts PI as input and achieve competitive performance.
We follow previous experiment setups~\citep{wang2023pi} and evaluated model performance on CIFAR-10H~\citep{peterson2019human} and CIFAR-10/100N~\citep{wei2021learning}.
These datasets involve relabeled versions of the original CIFAR.
For more details, see Appendix~\ref{app:pi}.
As a no-PI baseline, we train a Wide-ResNet-10-28~\citep{zagoruyko2016wide} model that ignores PI.
We also compare against recent noisy label learning methods: HET~\citep{collier2021correlated} and SOP~\citep{liu2022robust}.
We implement \MNAME{} with a similar architecture and training scheme as the no-PI baseline, where we concatenate the PI with the image-derived features and randomly knock PI out during training. 
For the common baseline, we train the same architecture with complete training data, but mean imputation for PI data during inference.
Table~\ref{tab:cifar10h_cifar10} lists test accuracy results.
For the CIFAR-10H dataset, where we have high quality PI, \MNAME~outperforms all baselines by a large margin, improving test accuracy by 6\%.
For CIFAR-10/100N datasets, where we have low quality PI during training, \MNAME's boost is more modest, performing similarly with SOP and slightly better than HET and the no-PI baseline.
\MNAME~can offer competitive results when we have access to high quality PI during training.

\begin{table}[t]
\centering
\caption{Average test accuracy (over 5 runs) of different methods on noisy label datasets with PI.
    For Quality, ``High'' and ``Low'' indicate access to sample-wise PI and batch-average PI respectively.
    Best results in \textbf{bold}, second-best \underline{underlined}.}\label{tab:cifar10h_cifar10}
    \begin{tabular}{l|c|rrr|r|r}
        \toprule
        Datasets          & Quality & No-PI               & HET                 & SOP                          & Common baseline             & \MNAME                       \\ \midrule
        CIFAR-10H (Worst) & High    & 51.1{\tiny$\pm$2.2} & 50.8{\tiny$\pm$1.4} & 51.3{\tiny$\pm$1.9}          & \underline{55.2{\tiny$\pm$0.8}} & \textbf{57.4{\tiny$\pm$0.6}} \\ \midrule
        CIFAR-10N (Worst) & Low     & 80.6{\tiny$\pm$0.2} & 81.9{\tiny$\pm$0.4} & \textbf{85.0\tiny{$\pm$0.8}} &       82.3\tiny{$\pm$0.3}              & \underline{84.7{\tiny$\pm$0.7}}          \\ \midrule
        CIFAR-100N (Fine) & Low     & 60.4{\tiny$\pm$0.5} & 60.8{\tiny$\pm$0.4} & \underline{61.9{\tiny$\pm$0.6}}          & 60.7{\tiny$\pm$0.6}                    & \textbf{62.1{\tiny$\pm$0.3}} \\ \bottomrule
    \end{tabular}
\end{table}

\subsection{Missing Images in Tumor Segmentation}\label{sec:multi_modal_tumor_segmentation}
\begin{figure}[t]
\centering
\includegraphics[width=\linewidth]{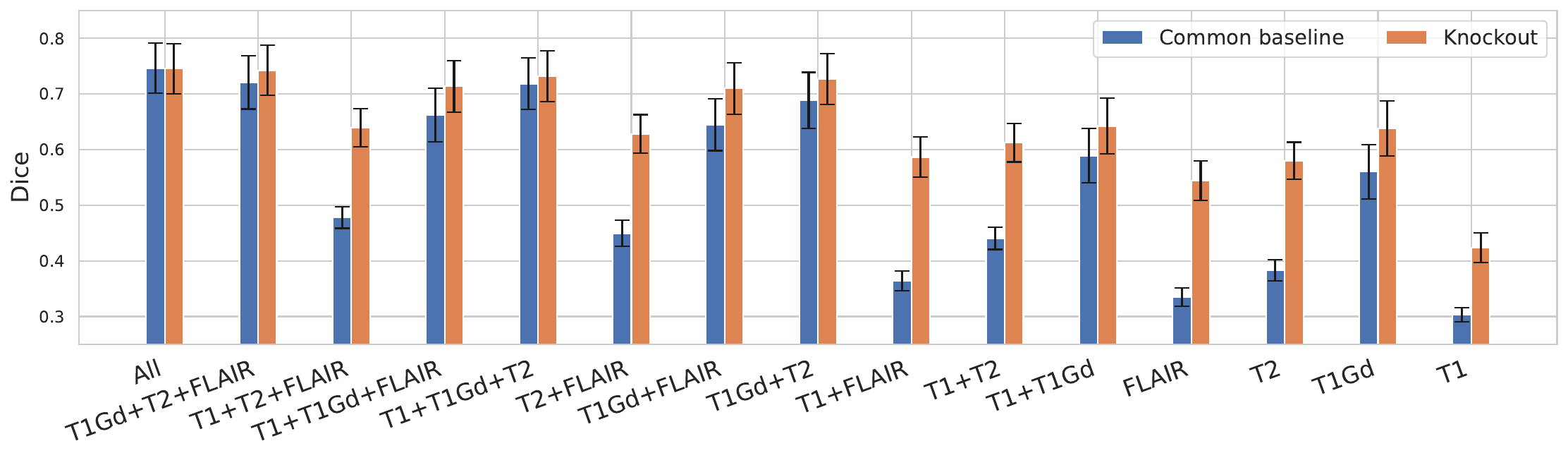}
    \caption{Dice performance of multi-modal tumor segmentation across varying missingness patterns of modality images.
    \MNAME~trained models have better Dice performance across all missingness patterns than the common baseline. 
    Error bars depict the 95\% confidence interval over test subjects.}\label{fig:lineplot_softdice_vs_dropoutp}
\end{figure}

Here, we investigate the ability of \MNAME~to handle missingness in a high-dimensional, 3D dense image segmentation task.
In particular, we experiment on a multi-modal tumor segmentation task~\citep{baid2021rsnaasnrmiccai}, where the goal is to delineate adult brain gliomas in 3D brain MRI volumes given 4 modalities per subject: T1, T1Gd, T2, and FLAIR.
We use a 3D UNet as the segmentation model~\citep{ronneberger2015unet}. 
We minimize a sum of cross-entropy loss and Dice loss with equal weighting and use Adam optimizer with a learning rate of 1e-3.
See Appendix~\ref{app:brats_dataset} and~\ref{sec:analysis_placeholders_structured_knockout} for further details.
At inference time, we evaluate on all modality missingness patterns.
Fig.~\ref{fig:lineplot_softdice_vs_dropoutp} shows Dice scores.
We observe that the \MNAME{}-trained model has better Dice performance across all missingness patterns.
When all modalities are available, \MNAME{} and the common baseline achieve the same performance level.
See Appendix~\ref{sec:analysis_placeholders_structured_knockout} for additional experiments on the effect of various placeholders.

\subsection{Missing Views in Tree Genus Classification}
We demonstrate \MNAME's ability to deal with missing data at the latent feature level in a classification task.
The Auto Arborist dataset~\citep{beery2022auto}, a multi-view (street and aerial) image dataset, is used for this purpose.
In this experiment, we used the top 10 genera for multi class prediction. 
A frozen ResNet-50~\citep{he2016deep} and ViT-B-16~\citep{dosovitskiy2020image} pretrained with ImageNet-v2~\citep{recht2019imagenet} is used as a feature extractor.
The two features from street and aerial images are concatenated and were fed into 3-layer MLP with ReLU activations.
We trained \MNAME~to randomly replace the whole latent vectors with vectors of 0s as placeholders after normalization.
This variant is denoted as \MNAME~(Structured).
We additionally trained two baselines for comparison: 1) \MNAME~(Features) where individual features in the latent vectors are independently replaced with placeholders, and 2) an imputation baseline, substituting latent vectors from missing views with vectors of zeros during inference.
Table~\ref{tbl:arborist-columbus} shows \MNAME~(Structured) outperforming \MNAME~(Features), suggesting that matching $p(\bm{M})$ with missing patterns that we expect to see at inference can be more effective.

\begin{table}[t]
\centering
\caption{F1-scores of Auto Arborist averaged over 5 random seeds. Each column represents non-missing modalities at inference time.
    Best results in \textbf{bold}, second-best \underline{underlined}.}
\begin{tabular}{ll rrr }
\toprule
&&Aerial+Street &Aerial &Street \\
\midrule
ResNet-50&Common baseline& 0.483$_{\pm 0.016}$ & \bf 0.312$_{\pm 0.017}$ & 0.356$_{\pm 0.024}$ \\
&\MNAME~(Features)     & \underline{0.493}$_{\pm 0.020}$ & 0.284$_{\pm 0.023}$ & \underline{0.381}$_{\pm 0.022}$ \\
&\MNAME~(Structured)   & \bf 0.496$_{\pm 0.016}$ & \underline{0.308}$_{\pm 0.024}$ & \bf 0.416$_{\pm 0.014}$ \\
\midrule
ViT-B-16&Common baseline & 0.464$_{\pm 0.018}$ & 0.305$_{\pm  0.022}$ & \underline{0.388}$_{\pm  0.011}$ \\
&\MNAME~(Features)     & \underline{0.473}$_{\pm 0.019}$ & \underline{0.315}$_{\pm  0.008}$ & 0.383$_{\pm  0.010}$ \\
&\MNAME~(Structured)   & \bf 0.480$_{\pm 0.017}$ & \bf 0.324$_{\pm  0.019}$ & \bf 0.408$_{\pm  0.015}$ \\
\bottomrule
\end{tabular}\label{tbl:arborist-columbus}
\end{table}

\subsection{Missing MR Modalities in Prostate Cancer Detection}
We demonstrate structured \MNAME~in the context of a binary image classification task, where \MNAME~is applied at the latent level.
The dataset consists of T2-weighted (T2w), diffusion-weighted (DWI) and apparent diffusion coefficient (ADC) MR images per subject~\citep{saha2022pi}.
A simple ``ensemble baseline'' approach to address missingness is to train a separate convolutional classifier for each modality, and average the predictions of available modalities at inference time~\citep{kim2023pulse,hu2020disentangled}.
For further details, please see~Appendix~\ref{app:prostate-cancer}.
To train a model with latent-level structured \MNAME, we use the same 3 feature extractors. 
Each feature extractor is trained with a different modality.
The loss function is binary cross entropy loss and we use an Adam optimizer with a learning rate of 1e-3.
We randomly knock out each modality. 
For the common baseline, we trained the same architecture with complete modalities.
At inference time, the latent features from missing modalities are imputed with 0s.
In the ``ensemble baseline'' approach, we averaged the predicted values from the three extractors without additional training.
As shown in Table~\ref{tab:prostate-f1}, \MNAME~generally outperforms the baselines in the majority of scenarios, except for inputs with ADC.
Notably, the F1 scores from the popular ensemble baseline are significantly lower than \MNAME.
\begin{table}[t]
\setlength{\tabcolsep}{2pt}
\centering
\caption{F1 scores of prostate cancer dataset averaged over 5 random seeds across varying missingness patterns at inference time.
    Each column represents non-missing modalities. Best results in \textbf{bold}, second-best \underline{underlined}.} 
\begin{tabular}{l rrr rrr r}
\toprule
&T2 &ADC &DWI &\makecell{ADC\\+DWI}&\makecell{T2\\+DWI}&\makecell{T2\\+ADC}&All\\
\midrule
Ensemble & 0.21\tiny{$\pm$0.09}&0.37\tiny{$\pm$0.01}&0.28\tiny{$\pm$0.03}&0.32\tiny{$\pm$0.01}&0.18\tiny{$\pm$0.04}&0.33\tiny{$\pm$0.03}&0.30\tiny{$\pm$0.05} \\
Common   & \underline{0.43}\tiny{$\pm$0.01}&\textbf{0.68}\tiny{$\pm$0.02}&\underline{0.61}\tiny{$\pm$0.02}&\textbf{0.70}\tiny{$\pm$0.01}&\underline{0.51}\tiny{$\pm$0.03}&\textbf{0.65}\tiny{$\pm$0.01}&\underline{0.67}\tiny{$\pm$0.01} \\
\MNAME   & \textbf{0.63}\tiny{$\pm$0.02}&\underline{0.60}\tiny{$\pm$0.02}&\textbf{0.62}\tiny{$\pm$0.02}&\underline{0.67}\tiny{$\pm$0.01}&\textbf{0.66}\tiny{$\pm$0.01}&\underline{0.65}\tiny{$\pm$0.02}&\textbf{0.68}\tiny{$\pm$0.01}\\
\bottomrule
\end{tabular}\label{tab:prostate-f1}
\end{table}

\subsection{Missing Modalities in UPMC Food-101 Classification}\label{ssec:food101}
To further evaluate the generalizability of our approach, we extend our experiments to the UPMC Food-101 dataset~\citep{wang2015recipe}, a widely used benchmark for multi-modal learning with both textual and visual inputs.
Our multi-modal architecture combines pretrained ResNet50 (vision) and RoBERTa-base (text) backbones.
The output from ResNet50 is fed into a single linear layer to transform its dimension to match that of Roberta’s output.
Feature vectors from each modality are concatenated and passed through an MLP with two hidden layers of size 256 and ReLU activations.
Models are trained using AdamW with a learning rate of $3 \times 10^{-5}$, for 8 epochs and a batch size of 72.
We follow the standard train-test split, reserving 7,000 samples from the training set for validation.

As shown in Table~\ref{tab:food101}, our multi-modal model with \MNAME~achieves 92.7\% accuracy, outperforming the common Baseline (91.5\%) and matching the previous state-of-the-art reported by~\citep{gallo2020image}.
Table~\ref{tab:food101} also reports performance under unimodal ablations.
\MNAME~exhibits greater robustness, especially in the image-only setting, improving accuracy from 47.5\% to 63.2\%.

\begin{table}[h]
\caption{Test accuracy (\%) on the Food-101 dataset for multi-modal models under three input settings: (1) both Image and Text, (2) Image only (without Text), and (3) Text only (without Image). For comparison, test accuracies of multi-modal models from previous work are also included.}\label{tab:food101}
\centering
\begin{tabular}{lccccc}
\toprule
Method & Image Model & Text Model & Image + Text  & w/o Text & w/o Image \\
\midrule
    \citet{wang2015recipe} & VGG19 & TF-IDF & 85.1 & N/A & N/A \\
    \citet{kiela2018efficient} & ResNet152 & FastText & 90.8 & N/A & N/A \\
    \citet{gallo2020image} & Inception-v3 & BERT-LSTM & 92.5 & N/A & N/A \\
\midrule
    Common baseline & ResNet50 & RoBERTa & 91.5 & 47.5 & 85.0 \\
    \MNAME          & ResNet50 & RoBERTa & 92.7 & 63.2 & 85.3 \\
\bottomrule
\end{tabular}
\end{table}

\section{Discussion}
We introduced \MNAME, a novel, easy-to-implement strategy for handling missing inputs, using a mathematically principled approach. 
By simulating missingness during training via random ``knock out'' and substitution with appropriate placeholder values, our method allows a single model to learn the conditional distribution and all desired marginals. 
Our evaluation demonstrates \MNAME's the versatility and robustness across diverse synthetic and real-world scenarios, consistently outperforming conventional imputation and ensemble techniques for both low and high-dimensional missing inputs.
We also extend \MNAME~to handle observed missing values in training and present a structured version that is more effective when entire feature vectors or input modalities are missing.

There has been a growing focus on developing multimodal datasets, tasks, and models that integrate information across modalities to improve performance and generalization~\cite{ngiam2011multimodal,srivastava2012multimodal,nguyen2018multimodal,lu2019vilbert,alayrac2022flamingo,boecking2022making}, motivating continued efforts to improve robustness to missing or incomplete inputs.
Although modern multimodal models are increasingly tolerant of missing modalities, \MNAME~can still play a crucial role in promoting robustness and adaptability.
This is especially useful during deployment, where models often face unpredictable or incomplete inputs due to sensor failures or missing data.
Multimodal foundation models like LLaVA~\citep{liu2023visual} and VILA~\citep{lin2024vila} are generally more robust to missing modalities than earlier models, but they are not immune to failure.
Their ability to handle missing inputs largely stems from pretraining on massive, diverse datasets.
However, when applied to tasks with input distributions that differ significantly from their training data, these models may still struggle in the absence of a modality.
Even SOTA transformer-based models have been shown to be sensitive to missing modalities~\citep{ma2022multimodal}.

There are several future directions for further investigation.
While our paper highlights the importance of choosing an appropriate placeholder value, and there appears to be a practical tension between selecting an unlikely/infeasible value versus achieving numerical stability (e.g., avoiding exploding gradients), one can conduct a more detailed study of this to optimize the placeholder value.
Additionally, our method assumes model training with SGD, where samples are seen multiple times.
Extending it to models that do not use SGD, such as XGBoost, is a valuable avenue.
Another promising direction of future research is adapting \MNAME~to address distribution shifts in the presence of missingness.
Finally, \MNAME's theoretical treatment hinges on the use of a high capacity, non-linear model trained on very large data. 
In applications, where low capacity models are used and/or training data are limited, \MNAME~might not be as effective.

\subsubsection*{Acknowledgments}
We thank Rachit Saluja, Leo Milecki, Haomiao Chen, and Benjamin C. Lee for the helpful comments.
Funding for this project was in part provided by the NIH grants R01AG053949, R01AG064027, R01AG070988 and 1K25CA283145, and the NSF CAREER 1748377 grant.

Data collection and sharing for this project was funded by the Alzheimer's Disease Neuroimaging Initiative
(ADNI) (National Institutes of Health Grant U01 AG024904) and DOD ADNI (Department of Defense award
number W81XWH-12-2-0012). ADNI is funded by the National Institute on Aging, the National Institute of
Biomedical Imaging and Bioengineering, and through generous contributions from the following: AbbVie,
Alzheimer’s Association; Alzheimer’s Drug Discovery Foundation; Araclon Biotech; BioClinica, Inc.; Biogen;
Bristol-Myers Squibb Company; CereSpir, Inc.; Cogstate; Eisai Inc.; Elan Pharmaceuticals, Inc.; Eli Lilly and
Company; EuroImmun; F. Hoffmann-La Roche Ltd and its affiliated company Genentech, Inc.; Fujirebio; GE
Healthcare; IXICO Ltd.; Janssen Alzheimer Immunotherapy Research \& Development, LLC.; Johnson \&
Johnson Pharmaceutical Research \& Development LLC.; Lumosity; Lundbeck; Merck \& Co., Inc.; Meso
Scale Diagnostics, LLC.; NeuroRx Research; Neurotrack Technologies; Novartis Pharmaceuticals
Corporation; Pfizer Inc.; Piramal Imaging; Servier; Takeda Pharmaceutical Company; and Transition
Therapeutics. The Canadian Institutes of Health Research is providing funds to support ADNI clinical sites
in Canada. Private sector contributions are facilitated by the Foundation for the National Institutes of Health
(www.fnih.org). The grantee organization is the Northern California Institute for Research and Education,
and the study is coordinated by the Alzheimer’s Therapeutic Research Institute at the University of Southern
California. ADNI data are disseminated by the Laboratory for Neuro Imaging at the University of Southern
California.

\bibliography{ref_cause}
\bibliographystyle{tmlr}

\clearpage
\appendix
\section{Appendix}
\subsection{Proof of Continuous and Unbounded Case}\label{app:cont_unbounded_proof}
Note that $p$ and $P$ denotes distribution and probability respectively.
As $P(X_i{=}\DEF_i)\approx 0$ implies $P(X_i{=}\DEF_i|\cdot)\approx 0$, as long as the conditioning is on a set of observations that is not extremely improbable. Then:
\begin{align} 
    P(X'_i{=}\DEF_i|\cdot) &= P(X'_i{=}\DEF_i,M_i=1|\cdot) + P(X'_i{=}\DEF_i,M_i=0|\cdot) \\
    &= P(M_i=1) P(X'_i{=}\DEF_i|M_i=1,\cdot) + P(M_i=0) P(X'_i{=}\DEF_i|M_i=0,\cdot) \\ 
    &= P(M_i=1) + P(M_i=0) P(X_i{=}\DEF_i|\cdot) \approx P(M_i=1) \\
    \Rightarrow p(Y|X'_i{=}\DEF_i,\cdot) &= \frac{p(Y|\cdot)P(X'_i{=}\DEF_i|Y,\cdot)}{P(X'_i{=}\DEF_i|\cdot)} \approx \frac{p(Y|\cdot)P(M_i=1)}{P(M_i=1)} \approx p(Y|\cdot) \label{eq:approx_condition}
    \end{align}
A similar analysis can be performed to derive the approximate versions of equations of Eq.~\eqref{eq:first_in_set} to~\eqref{eq:last_in_set}, which are special cases.

\subsection{Suboptimal Choice of Placeholder Values}\label{app:counterexample}

If $\DEF_i \in \text{support}(X_i)$, then $X_i'=\DEF_i$ is either because of two mutually exclusive cases:
\[ \begin{cases}
    M_i=1 \text{ or} \\
    M_i=0 \text{ and } X_i=\DEF_i
\end{cases} \]
Let $P(M_i=0)=r$ and $\bm{x}_{j} \neq \DEF_{j}, \forall j \neq i$.
Assuming that $P(X_i'{=}\DEF_i|\bm{X}_{-i}{=}\bm{x}_{-i}) > 0$, then:
\begin{align}
    p(Y|X_i'{=}\DEF_i,\bm{X}_{-i}'{=}\bm{x}_{-i}) &= p(Y|X_i'{=}\DEF_i,\bm{X}_{-i}{=}\bm{x}_{-i}) = \frac{p(Y,X_i'{=}\DEF_i|\bm{X}_{-i}{=}\bm{x}_{-i})} {P(X_i'{=}\DEF_i|\bm{X}_{-i}{=}\bm{x}_{-i})} \\
    P(X_i'{=}\DEF_i|\bm{X}_{-i}{=}\bm{x}_{-i}) &= P(M_i=1) + P(M_i=0,X_i{=}\DEF_i|\bm{X}_{-i}{=}\bm{x}_{-i}) \\
    &= 1-r + P(X_i{=}\DEF_i|M_i=0,\bm{X}_{-i}{=}\bm{x}_{-i})P(M_i=0|\bm{X}_{-i}{=}\bm{x}_{-i}) \\
    &= 1-r + r\times P(X_i{=}\DEF_i|\bm{X}_{-i}{=}\bm{x}_{-i}) \label{eq:denom} \\
    p(Y,X_i'{=}\DEF_i|\bm{X}_{-i}{=}\bm{x}_{-i}) &= p(Y|\bm{X}_{-i}{=}\bm{x}_{-i}) P(X_i'{=}\DEF_i|Y,\bm{X}_{-i}{=}\bm{x}_{-i}) \\
    &= p(Y|\bm{X}_{-i}{=}\bm{x}_{-i}) (1-r + r\times P(X_i{=}\DEF_i|Y,\bm{X}_{-i}{=}\bm{x}_{-i})) \label{eq:nom}
\end{align}
From Eq.~\eqref{eq:denom} and~\eqref{eq:nom},
\begin{equation}
    p(Y|X_i'{=}\DEF_i,\bm{X}_{-i}'{=}\bm{x}_{-i}) = p(Y|\bm{X}_{-i}{=}\bm{x}_{-i}) \frac{1-r + r\times P(X_i{=}\DEF_i|Y,\bm{X}_{-i}{=}\bm{x}_{-i})} {1-r + r\times P(X_i{=}\DEF_i|\bm{X}_{-i}{=}\bm{x}_{-i})} \\
\end{equation}
Thus, $p(Y|X_i'{=}\DEF_i,\bm{X}_{-i}'{=}\bm{x}_{-i}) = p(Y|\bm{X}_{-i}{=}\bm{x}_{-i})$ is equivalent to
\begin{align}
    &\Leftrightarrow 1-r + r\times P(X_i{=}\DEF_i|Y,\bm{X}_{-i}{=}\bm{x}_{-i}) = 1-r + r\times P(X_i{=}\DEF_i|\bm{X}_{-i}{=}\bm{x}_{-i}) \\
    &\Leftrightarrow P(X_i{=}\DEF_i|Y,\bm{X}_{-i}{=}\bm{x}_{-i}) = P(X_i{=}\DEF_i|\bm{X}_{-i}{=}\bm{x}_{-i}) \\
    &\Leftrightarrow Y\indep X_i{=}\DEF_i | \bm{X}_{-i}{=}\bm{x}_{-i} \label{eq:impossible_criterion}
\end{align}
It is difficult to find $\DEF_i$ to satisfy Eq.~\ref{eq:impossible_criterion}, most likely $p(Y|X_i'{=}\DEF_i,\bm{X}_{-i}'{=}\bm{x}_{-i})\neq p(Y|\bm{X}_{-i}{=}\bm{x}_{-i})$.

For example, consider the case where $E_X=\{x_1, x_2\}, E_Y=\{0, 1\}, x_0:=x_1$, and $\alpha=0.5$.
\begin{align}
    &\PP(X=x_1) = 0.3,\quad \PP(X=x_2) = 0.7 \\
    &\PP(Y=0|X=x_1) = \PP(Y=1|X=x_2) = 1 \\
    &\Rightarrow \PP(Y=0) = 0.3,\quad \PP(Y=1) = 0.7 \\
    &\PP(Y=0|X'=x_1) = \frac{\PP(Y=0,X'=x_1)}{\PP(X'=x_1)} \\
    &= \frac{\PP(Y=0,M_X=1) + \PP(Y=0,M_X=0,X=x_1)}{\PP(M_X=1) + \PP(M_X=0,X'=x_1)} \\
    &= \frac{0.3*0.5 + 0.5*\PP(X=x_1)\PP(Y=0|X=x_1)}{0.5 + 0.5*0.3} \\
    &= \frac{0.3*0.5 + 0.5*0.3*1}{0.5 + 0.5*0.3} = \frac{2*0.3*0.5}{0.5 + 0.5*0.3} = \frac{0.6}{1.3} \\
    &\Rightarrow \PP(Y=0|X'=x_1) \neq \PP(Y=0) = 0.3 \\
    &\Rightarrow \PP(Y=0|X'=x_1) \neq \PP(Y=0|X=x_1) = 1
\end{align}

\subsection{Analysis of Placeholders for Structured \MNAME}\label{sec:analysis_placeholders_structured_knockout}
In this subsection, we analyze the effect of different placeholder values on a structured \MNAME{} task, specifically the multi-modal tumor segmentation task from Section~\ref{sec:multi_modal_tumor_segmentation}.
We scale image intensity values to $[0, 1]$ per image.
We train three \MNAME~models with the following placeholders: a constant image of -1s, a constant image of 0s, and the mean of all images per modality.
At inference time, we evaluate on all modality missingness patterns.
In the event that all images are missing, we randomly select one so that the model sees at least one image.
For \MNAME-trained models, the corresponding placeholder is imputed for missing images.

\begin{figure}[h]
\centering
\includegraphics[width=\linewidth]{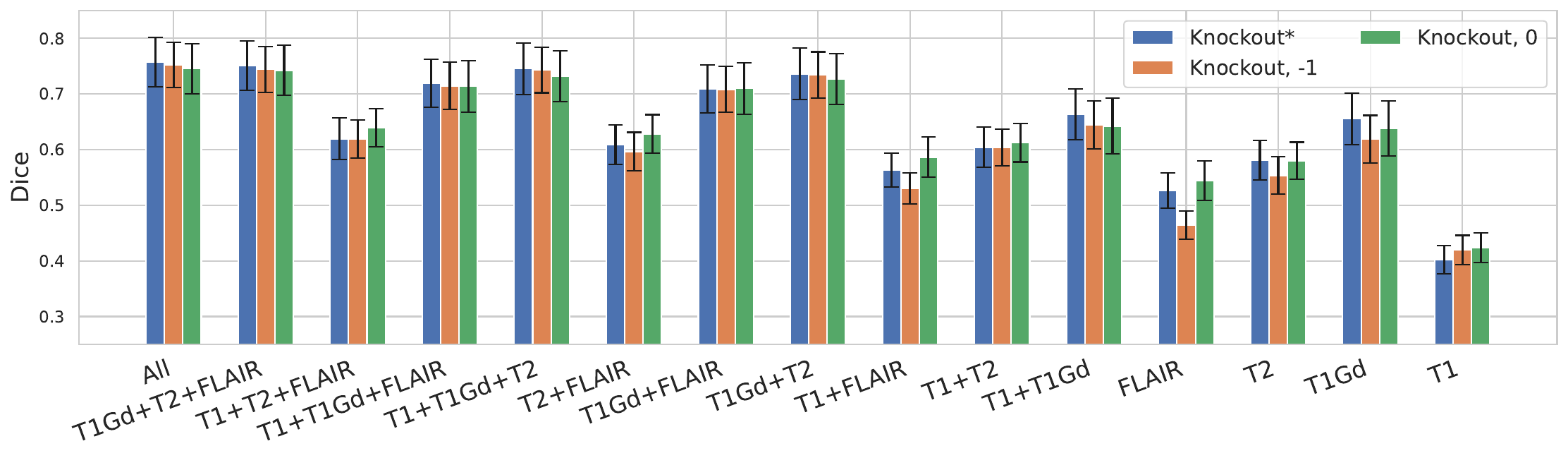}
    \caption{Dice performance of multi-modal tumor segmentation across varying missingness patterns of modality images. \MNAME{}-trained models only. 
    We observe mean placeholders perform better than constant-image placeholders. Error bars depict the 95\% confidence interval over test subjects.
    }\label{fig:barplot_softdice_vs_every_missingness_knockoutonly}
\end{figure}

Fig.~\ref{fig:barplot_softdice_vs_every_missingness_knockoutonly} shows the results.
Interestingly, we observe the mean placeholder (\MNAME*) performs better than constant-image placeholders, and the constant image of 0s generally outperforms the constant image of -1s.
We hypothesize that in the context of structured inputs like images in conjunction with limited data and model capacity, placeholders which balance feasibility with practical considerations like causing unstable gradients due to out-of-range inputs is an important consideration.

\section{Experimental Details}
All experiments were performed with access to a machine equipped with an AMD EPYC 7513 32-Core processor and an Nvidia A100 GPU.
All code is written in PyTorch.  

\subsection{Simulations}\label{app:simulations}

\subsubsection{Regression}\label{app:regression}
In each repetition, the data are sampled from a 10-dimensional multivariate Gaussian distribution with mean $\bm{\mu}$ and covariance $\bm{\Sigma}$.
The mean vector $\bm{\mu}$ is sampled uniformly from the interval $[0, 1]$, i.e.~$\bm{\mu}\sim\mathsf{Uniform}(0, 1)\in\mathbb{R}^{10}$.
The covariance matrix is sampled as $\bm{\Sigma}:=\bm{W}^T \bm{W}$, whereby $\bm{W}\sim\mathsf{Uniform}(0, 1)\in\mathbb{R}^{10\times 10}$.
The full training and test datasets are then generated using the covariance matrix (fixed).
This ensures consistency and avoids introducing bias that could affect predictive imputation methods.
The first 9th variables of the multivariate Gaussian are assigned as $\bm{X}$ ($\bm{X}\in\mathbb{R}^9$) and the 10th variable is assigned as $Y$ ($Y\in\mathbb{R}$).
The Knockout rate is set at 0.0741 so that half of the mini-batches have no induced missing variable.

\begin{figure}[t]
\centering
\includegraphics[width=\linewidth]{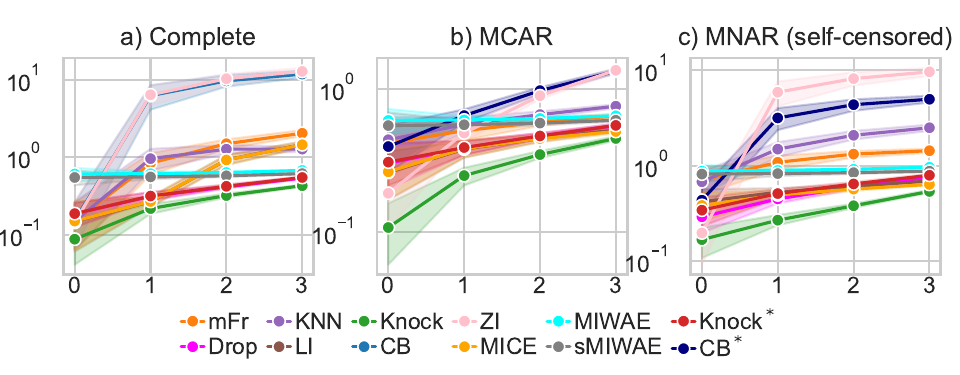}
    \caption{Test MSE evaluated against observations ($Y$) from 10 repetitions of the regression simulation.
    Lower is better.
    X axis indicates the number of missing variables at inference time.
    a) Complete training data.
    b) Missing completely at random (MCAR) training data.
    c) Missing not at random (MNAR) training data}\label{fig:sim1_gt}
\end{figure}
\begin{figure}[t]
\centering
\includegraphics[width=\linewidth]{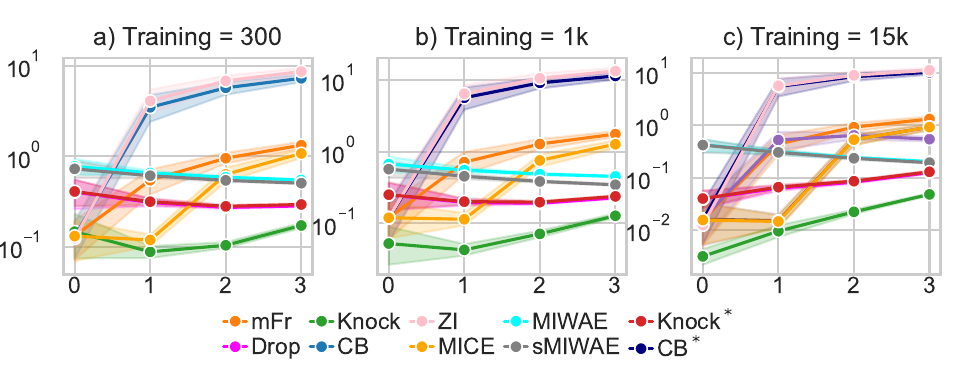}
    \caption{\MNAME~at low-data and high-data settings.
    Test MSE evaluated against Bayes optimal prediction ($\EE[Y|\bm{X}]$) averaged over 10 repetitions (lower is better).
    X axis indicates the number of missing variables at inference time.}\label{fig:sim_size}
\end{figure}

In addition to the MMSE-minimizing Bayes optimal predictions: $\EE[Y | \bm{X}]$, we also evaluate the models' predictions against the observed values of $Y$ (Fig.~\ref{fig:sim1_gt}).
To assess model robustness across varying data availability, we experiment with training set sizes of 300, 1000, and 15000 samples (compared to the original 3000-sample setting).
This ablation focuses exclusively on fully observed training data.
As illustrated in Fig.~\ref{fig:sim_size}, \MNAME~consistently outperforms all baseline methods across all levels of missingness.
With only 300 samples, \MNAME~achieves the lowest mean squared error (MSE) at every level of missingness.
This performance trend continues at 1000 samples, with \MNAME~again leading.
Although all methods benefit from the increased training data at 15000 samples, \MNAME~maintains a clear and consistent performance advantage.
These results underscore \MNAME's robustness and strong generalization, especially under high missingness and limited data conditions.

\subsubsection{Binary Classification}\label{app:bin_classification}
\begin{figure}[h]
    \centering
    \begin{subfigure}{0.48\linewidth}
        \centering
        \includegraphics[width=\linewidth]{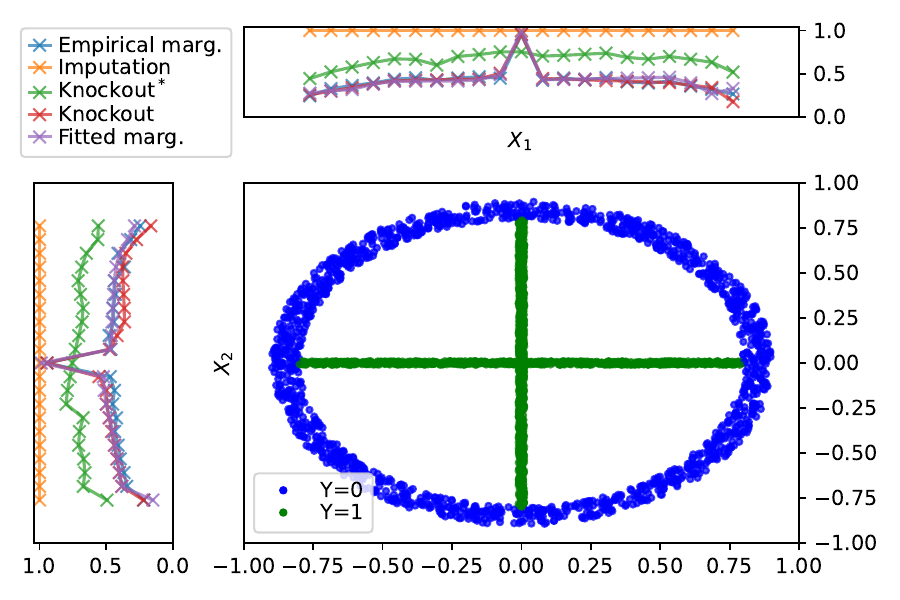}
        \caption{Continuous inputs}\label{fig:sim2}
    \end{subfigure}
    \hfill
    \begin{subfigure}{0.48\linewidth}
        \centering
        \includegraphics[width=\linewidth]{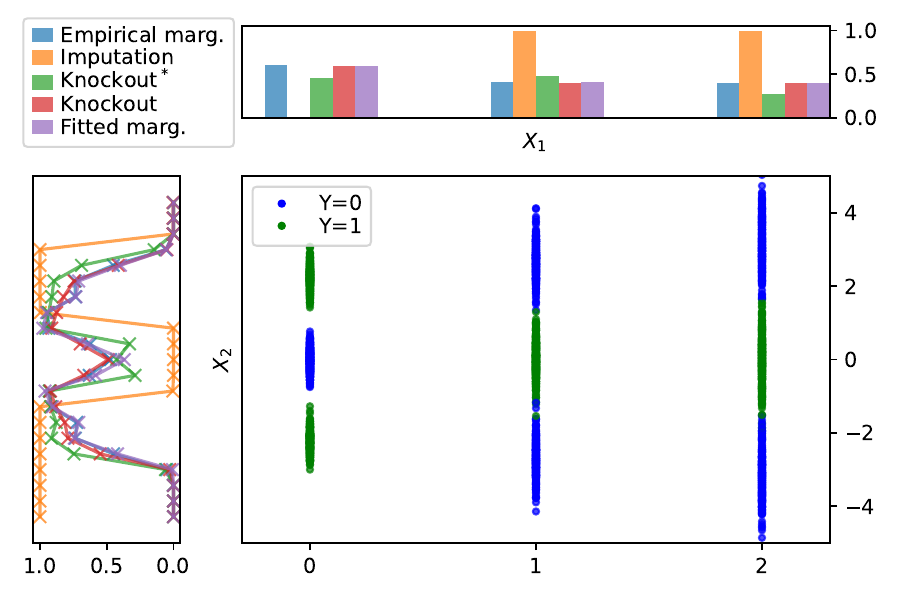}
        \caption{Mixed inputs}\label{fig:sim3}
    \end{subfigure}
\caption{Visualization of the two classification simulations.
    \MNAME's estimates of the marginal distributions (i.e.~$P(Y|X_1)$ and $P(Y|X_2)$, denoted by red lines) are closer to the empirical estimates (blue lines) than baselines'.
    Top: $P(Y{=}1|X_1)$ estimated empirically and estimated by various approaches.
    Left: Various estimates of $P(Y{=}1|X_2)$.
    Bottom Right: Data visualization.
    }\label{}
\end{figure}

We evaluate the prediction error rate with full features ($\bm{X}$) and missing feature (only $X_1$ or $X_2$ as input).
We also evaluate how close the models' predicted probability distributions with missing feature are against the marginal distributions (Fig.~\ref{fig:sim2} and Fig.~\ref{fig:sim2} top and left panels) using Jensen-Shannon divergence.
The marginals distributions are estimated empirically using all data.

\textbf{Continuous Inputs.}
All input variables are continuous ($\bm{X}\in\mathbb{R}^2$).
\MNAME~achieves similar error rate compared to standard training but much better performance when a variable is missing (Table~\ref{tbl:sim23}).
\MNAME$^*$ performs worse than \MNAME~due to the sub-optimal choice of placeholders.

\begin{table}
  \caption{Classification simulations. Best results are in bold.
  Err.: Proportion of test error. JSD: Jensen–Shannon divergence of the estimated and empirical marginal.}\label{tbl:sim23}
  \centering
  \begin{tabular}{lccccc}
    \toprule
      & Missing Rate = 0 & \multicolumn{4}{c}{Missing Rate = 1/2} \\ %
    \cmidrule(r){3-6}
      Method        & Err. ($\bm{X}$) & Err. ($X_1$) & JSD ($X_1$) & Err. ($X_2$) & JSD ($X_2$) \\ %
    \midrule
    \multicolumn{6}{c}{Continuous inputs} \\
    \midrule
      Common baseline    & \bf 0.0003 &     0.3970 &   $\infty$ &     0.4001 &   $\infty$ \\ %
      \MNAME$^*$~(Ours)  &     0.0210 &     0.3549 &     0.0179 &     0.3727 &     0.0214 \\ %
      \MNAME~(Ours) &     0.0007 & \bf 0.2559 & \bf 0.0003 & \bf 0.2563 & \bf 0.0007 \\ %
    \midrule
      Fitted Marginals   &     N/A    &     0.2531 &     0.0006 &     0.2600 &     0.0006 \\ %
    \midrule
    \multicolumn{6}{c}{Mixed inputs} \\
    \midrule
      Common baseline    &     0.0032 &     0.5972 &   $\infty$ &     0.5410 &   $\infty$ \\ %
      Common baseline ($\DEF$) &  0.0032 &     0.4843 &   $\infty$ &     0.4137 &   $\infty$ \\ %
      \MNAME$^*$~(Ours)            &     0.0187 &     0.4843 &     0.0038 &     0.3410 &     0.0073 \\ %
      \MNAME~(Ours) & \bf 0.0031 & \bf 0.4028 & \bf 0.0001 & \bf 0.2844 & \bf 0.0009 \\ %
    \midrule
      Fitted Marginals            &     N/A    &     0.4028 &     0.0000 &     0.2809 &     0.0008 \\ %
    \bottomrule
  \end{tabular}
\end{table}

\textbf{Mixed Inputs.}
$\bm{X}$ consists of a binary variable and a continuous variable (i.e.~$X_1\in\{0,1\},\;\;X_2\in\mathbb{R}$).
We also include a common baseline (CB) approach that imputes using $\DEF$.
CB ($\DEF$) is similar to CB but impute missing variables using new categories and out-of-support values during inference.
Since CB ($\DEF$) lacks a learned representation for these values, its performance is less reliable and may not generalize well.
In contrast, \MNAME~offers greater robustness by explicitly modeling missingness during training.
\MNAME~achieves better results than baselines in all scenarios (Table~\ref{tbl:sim23}).

\subsection{Alzheimer's Forecasting}\label{app:ad}
All participants used in this work are from the Alzheimer's Disease Neuroimaging Initiative (ADNI) database (adni.loni.usc.edu).
The ADNI was launched in 2003 as a public-private
partnership, led by Principal Investigator Michael W. Weiner, MD. The primary goal of ADNI has been to
test whether serial magnetic resonance imaging (MRI), positron emission tomography (PET), other
biological markers, and clinical and neuropsychological assessment can be combined to measure the
progression of mild cognitive impairment (MCI) and early Alzheimer’s disease (AD).

We select the participants who have mild cognitive impairment (MCI) at the baseline (screening) visit and had at least one follow-up diagnosis within the next five years. 
We excluded participants who were diagnosed as CN in a later follow-up year (n=284) since these subjects might have been diagnosed incorrectly at some point.
After this exclusion, we are left with 789 participants. 
Table~\ref{table:ad1} lists summary statistics for the participants; including sex, age, number of years of education completed, count of Apolipoprotein E4 (APOE4) allele, Clinical Dementia Rating(CDR), and Mini Mental State Examination (MMSE) scores at baseline.
As is common in many real-world longitudinal studies, ADNI experiences missing follow-up visits, irregular timings, and high dropout rates before the study's planned end. 
Table~\ref{table:ad2} shows the number of subjects available in each diagnostic category for annual follow-ups. 
In Table~\ref{table:ad2} and all analyses, any subject who progressed from MCI to AD before withdrawing was considered to remain in the AD state until the fifth year, reflecting AD's irreversible nature.
We employed the reweighted cross-entropy loss scheme introduced in \citep{karaman_2022_machine} during training to account for the imbalance in diagnoses.

\begin{table}
\centering
\caption{Summary statistics of the participants at baseline in the Alzheimer's Disease data. Mean $\pm$ standard deviations are listed. APOE4 row represents the number of alleles. }\label{table:ad1}
\centering
\begin{tabular}{ll}
\toprule
Characteristic & ($n$=789) \\
\midrule
Female/Male & \ensuremath{324/465} \\ 
Age $(yr)$ & \ensuremath{73.46 \pm 7.39} \\ 
Education $(yr)$ & \ensuremath{15.93 \pm 2.81} \\ 
APOE4 $(0/1/2)$ & \ensuremath{371/313/98} \\ 
CDR & \ensuremath{1.55 \pm 0.89} \\ 
MMSE & \ensuremath{27.52 \pm 1.82} \\ 
\bottomrule
\end{tabular}
\end{table}

\begin{table}
\centering
\caption{The number of available subjects in each diagnostic group for annual follow-up visits in the Alzheimer's Disease data. The follow-up diagnoses are not actually exactly 12 months apart. They have been rounded to the nearest time horizon in years.}\label{table:ad2}
\begin{tabular}{llllll}
\toprule
Follow-up year & 1 & 2 & 3 & 4 & 5\\ 
\midrule
MCI & 674 & 431 & 317 & 202 & 127\\ 
AD  & 110 & 218 & 261 & 286 & 292\\ 
\bottomrule
\end{tabular}
\end{table}

We use features from several domains.
Demographics include age and years of education (PTEDUCAT), and genotype is represented by APOE4 allele count.
Cognitive assessments include CDR, FAQ, ADAS (11, 13, Q4), MMSE, RAVLT, LDELTOTAL, TRABSCOR, mPACCdigit and mPACCtrailsB. 
Biomarkers include CSF measures (ABETA, TAU, PTAU), MRI-derived brain volumes (ventricles, hippocampus, whole brain, entorhinal, fusiform, mid-temporal, and ICV) from FreeSurfer~\citep{Desikan2006968,Fischl01012004}, and PET SUVR for the FDG tracer.
All features are numerical and z-score normalized using training-set statistics.
Missing rates per modality are shown in Table~\ref{table:ad3}.
\begin{table}
\centering
\caption{The degree of missingness ($\%$) in different data modalities in the Alzheimer's Disease data.}\label{table:ad3}
\begin{tabular}{ll}
\toprule
Data Type & Missingness Rate (\%)\\ 
\midrule
Demographics & 0.06\\
Genotype & 0.89\\
Cognitive assessments & 0.20\\ 
CSF & 37.14\\ 
MRI  & 21.22\\ 
FDG & 24.08\\ 
\bottomrule
\end{tabular}
\end{table}

\begin{figure}[h]
\centering
\includegraphics[width=\linewidth]{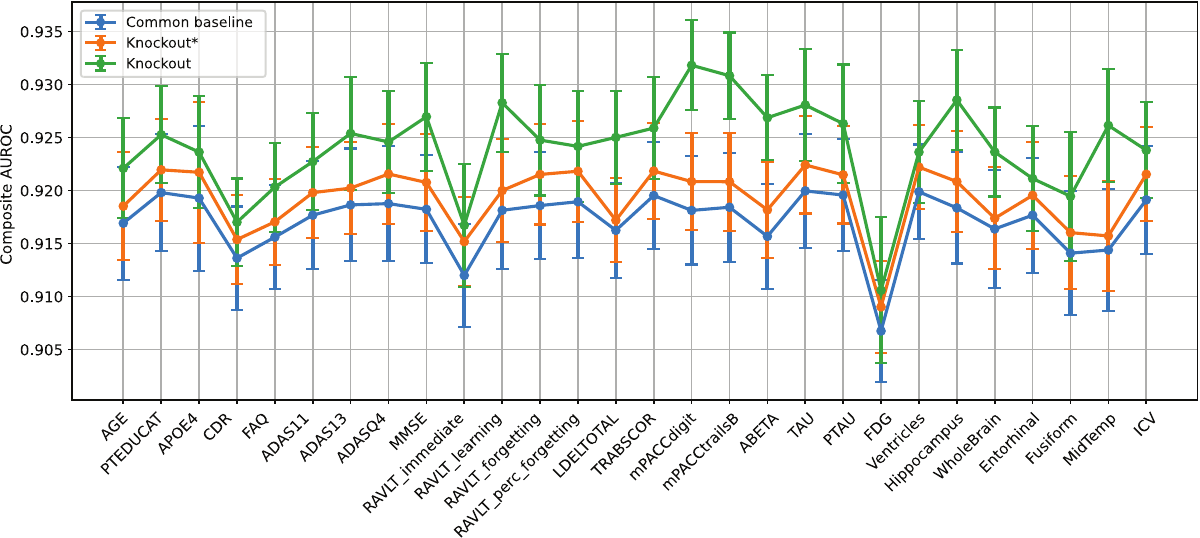}
\caption{AUROC scores obtained for the three model variants when each input feature is missing during inference (x-axis) for the complete data case in the Alzheimer's Disease forecasting experiment. Displayed are averages of 10 train-test splits. Error bars indicate the standard error across these splits.
}\label{fig:ad-supp} 
\end{figure}

We note that we train our models using the hyperparameters stated in \citep{karaman_2022_machine}.
Fig.~\ref{fig:ad-supp} shows the AUROC scores obtained using the complete portion of the dataset ($n=256$ subjects).
In this experiment, the training data has no observed missing variables.
These results are similar to the results included in the main text, where \MNAME~outperforms the baseline and the choice of the appropriate placeholder has an impact on the performance.

\subsection{Multi-modal Tumor Segmentation}\label{app:brats_dataset}
The RSNA-ASNR-MICCAI BraTS~\citep{baid2021rsnaasnrmiccai} challenge releases a dataset of 1251 subjects with multi-institutional routine clinically-acquired multi-parametric MRI scans of glioma.
Each subject has 4 modalities: native (T1), post-contrast T1-weighted (T1Gd), T2-weighted (T2), and T2 Fluid Attenuated Inversion Recovery (T2-FLAIR).
All the imaging datasets have been annotated manually, by one to four raters, following the same annotation protocol, and their annotations were approved by experienced neuro-radiologists. Annotations comprise the GD-enhancing tumor (ET - label 3), the peritumoral edematous/invaded tissue (ED - label 2), and the necrotic tumor core (NCR - label 1). 

The following pre-processing is applied: co-registration to the same anatomical template, interpolation to the same resolution (1 mm3), skull-stripped, and min-max normalized to the range [0, 1].
The ground truth data were created after their pre-processing.
For training, we use 80\%/5\%/15\% data split of the subjects for training/validation/testing.

For the segmentation model, we use a 3D UNet with 4 downsampling layers and 2 convolutional blocks per resolution~\citep{ronneberger2015unet}. 
We minimize a sum of cross-entropy loss and Dice loss with equal weighting and use an Adam optimizer with a learning rate of 1e-3.

\subsection{Prostate Cancer Detection}\label{app:prostate-cancer}
A common clinical workflow for the diagnosis of prostate cancer is to detect and localize abnormalities from 3 MR modalities: T2-weighted (T2w), diffusion-weighted (DWI) and apparent diffusion coefficient (ADC) images~\citep{turkbey2019prostate}.
T2w images provide anatomical details, while DWI and ADC highlight restricted diffusion, which can be a sign of malignancy. 

We divided 1500 biparametric MR image sets provided from Prostate Imaging: Cancer AI (PICAI) challenge~\citep{saha2022pi} "training" dataset into training, validation, test sets in a 0.6/0.2/0.2 ratio. Among the 1500 cases, 425 were confirmed as cancer by biopsy. DWI and ADC images are registered to T2w images and all images are cropped around prostate and resized to $100\times100\times40$. For the modality-wise classification tasks, we used 3D CNN with 4 blocks, each with a convolution layer, BatchNorm, leakly ReLU activation and average pooling layer, followed by fully connected layer. We trained the models to predict PCa using binary cross entropy loss and an Adam optimizer with a learning rate of $1e-3$. 

\begin{table}[h!]
\setlength{\tabcolsep}{3pt}
\centering
\caption{AUC performance for prostate cancer detection from the ensemble baseline, common baseline, and {\MNAME}, across varying missingness patterns at inference time. Each column represents non-missing modalities. Best results in \textbf{bold}, second-best \underline{underlined}.} 
\begin{tabular}{l ccc ccc c}
\toprule
&T2 &ADC &DWI &\makecell{ADC\\+DWI}&\makecell{T2\\+DWI}&\makecell{T2\\+ADC}&All\\
\midrule
Ensemble   &0.683\tiny{$\pm$0.013}&\textbf{0.786}\tiny{$\pm$0.010}&0.718\tiny{$\pm$0.007}&0.780\tiny{$\pm$0.005}&0.722\tiny{$\pm$0.005}&\underline{0.766}\tiny{$\pm$0.006}&\underline{0.766}\tiny{$\pm$0.004}\\
Common   &\underline{0.687}\tiny{$\pm$0.011}&\underline{0.771}\tiny{$\pm$0.011}&\underline{0.720}\tiny{$\pm$0.007}&\underline{0.784}\tiny{$\pm$0.003}&\underline{0.727}\tiny{$\pm$0.006}&\textbf{0.771}\tiny{$\pm$0.008}&\textbf{0.774}\tiny{$\pm$0.004}\\
{\MNAME}    &\textbf{0.694}\tiny{$\pm$0.009}&{0.730}\tiny{$\pm$0.019}&\textbf{0.736}\tiny{$\pm$0.009}&\textbf{0.789}\tiny{$\pm$0.005}&\textbf{0.744}\tiny{$\pm$0.004}&{0.753}\tiny{$\pm$0.011}&\textbf{0.774}\tiny{$\pm$0.007}\\
\bottomrule
\end{tabular}\label{tab:prostate-auc}
\end{table}

\subsection{Privileged information for noisy label learning}\label{app:pi}
We briefly introduce two datasets we used for this experiment: CIFAR-10H~\citep{peterson2019human} and CIFAR-10/100N~\citep{wei2021learning}.
CIFAR-10H relabels the original CIFAR-10 10K test set with multiple annotators and provides high-quality sample-wise annotation information such as annotator ID, reaction time and annotator confidence as PI.
Following a previous setup~\citep{wang2023pi}, we test on the high-noise version of CIFAR-10H, by selecting incorrect labels when available, denoted as "CIFAR-10H Worst".
The estimated noise rate is 64.6\%.
While we train on the high-noise version, testing is conducted on the original CIFAR-10 50K training set.
CIFAR-10/100N provides multiple annotations for CIFAR-10/100 training set.
The raw data also includes information about annotation process.
But this information is provided as averages over batches of examples rather than sample-wise.
The estimated noise rate is 40.2\% for CIFAR-10/100N. 

For all CIFAR experiments and baselines, we use the Wide-ResNet-10-28~\citep{zagoruyko2016wide} architecture.
We use SGD optimizer with 0.9 Nesterov momentum, a batch size of 256, 0.1 learning rate and 1e-3 weight decay and minimized the cross-entropy loss with respect to the provided labels.
The total training epoch is 90, and the learning rate decayed by a factor of 0.2 after 36, 72 epochs.
For the PI features, we use annotator ID and annotation reaction time.
In PI features are normalized to [0, 1] for preprocessing.
For \MNAME, during training, we randomly knock out all PI features at $50\%$ rate and use -1 as placeholder value.
All experiments are performed on one A6000.
In Table~\ref{tab:pi_knockoutstar}, we further show results for common baseline, \MNAME$^*$ and \MNAME.

\begin{table}[]
\centering
\begin{tabular}{l|c|c|c|c|c}
\toprule 
Datasets          & PI quality & SOP                             & Common baseline               & \MNAME$^*$                  & \MNAME                       \\ \hline
CIFAR-10H (Worst) & High       & 51.3{\tiny$\pm$1.9}             & 55.2\tiny$\pm$0.8 & \underline{56.9\tiny{$\pm$0.59}}    & \textbf{57.4{\tiny$\pm$0.6}} \\ \hline
CIFAR-10N (Worst) & Low        & \textbf{85.0\tiny{$\pm$0.8}}    & 82.3\tiny{$\pm$0.3}           &   83.6\tiny$\pm$0.7                        & \underline{84.7\tiny$\pm$0.7}  \\ \hline
CIFAR-100N (Fine) & Low        & 61.9\tiny$\pm$0.6 & 60.7\tiny$\pm$0.6           &  \underline{61.6\tiny$\pm$0.6} & \textbf{62.1\tiny$\pm$0.3} \\ \bottomrule
\end{tabular}
\caption{Test accuracy of different methods on noisy label dataset with PI. Best results in \textbf{bold}, second-best \underline{underlined}.}\label{tab:pi_knockoutstar}
\end{table}

\end{document}